\documentclass[10pt,twocolumn,letterpaper]{article}

\usepackage[pagenumbers]{cvpr} 
\usepackage{multirow}
\definecolor{cvprblue}{rgb}{0.21,0.49,0.74}
\usepackage[pagebackref,breaklinks,colorlinks,allcolors=cvprblue]{hyperref}

\def\paperID{*****} 
\def\confName{CVPR}
\def\confYear{2026}

\title{Pi-HOC: Pairwise 3D Human-Object Contact Estimation}

\author{
  Sravan Chittupalli$^{1,2}$ \quad Ayush Jain$^{1}$ \quad Dong Huang$^{1,2}$\\
  $^{1}$Carnegie Mellon University, Robotics Institute\\
  $^{2}$National Robotics Engineering Center\\
  {\tt\small \{schittup,ayushj2,dghuang\}@andrew.cmu.edu}
}

\begin{document}
\maketitle

\begin{abstract}

Resolving real-world human-object interactions in images is a many-to-many challenge, in which disentangling fine-grained concurrent physical contact is particularly difficult. Existing semantic contact estimation methods are either limited to single-human settings or require object geometries (e.g., meshes) in addition to the input image. Current state-of-the-art leverages powerful VLM for category-level semantics but struggles with multi-human scenarios and scales poorly in inference. We introduce \textbf{Pi-HOC, a single-pass, instance-aware framework for dense 3D semantic contact prediction of all human-object pairs}. Pi-HOC detects instances, creates dedicated human-object (HO) tokens for each pair, and refines them using an InteractionFormer. A SAM-based decoder then predicts dense contact on SMPL human meshes for each human-object pair. On the MMHOI and DAMON datasets, Pi-HOC significantly improves accuracy and localization over state-of-the-art methods while achieving $20\times$ higher throughput. We further demonstrate that predicted contacts improve SAM-3D image-to-mesh reconstruction via a test-time optimization algorithm and enable referential contact prediction from language queries without additional training. Our code and checkpoints  can be found at the project website  \url{https://pi-hoc.github.io/}. 
\end{abstract}

\section{Introduction}

\begin{figure*}[t]
  \begin{center}
    \includegraphics[width=\textwidth]{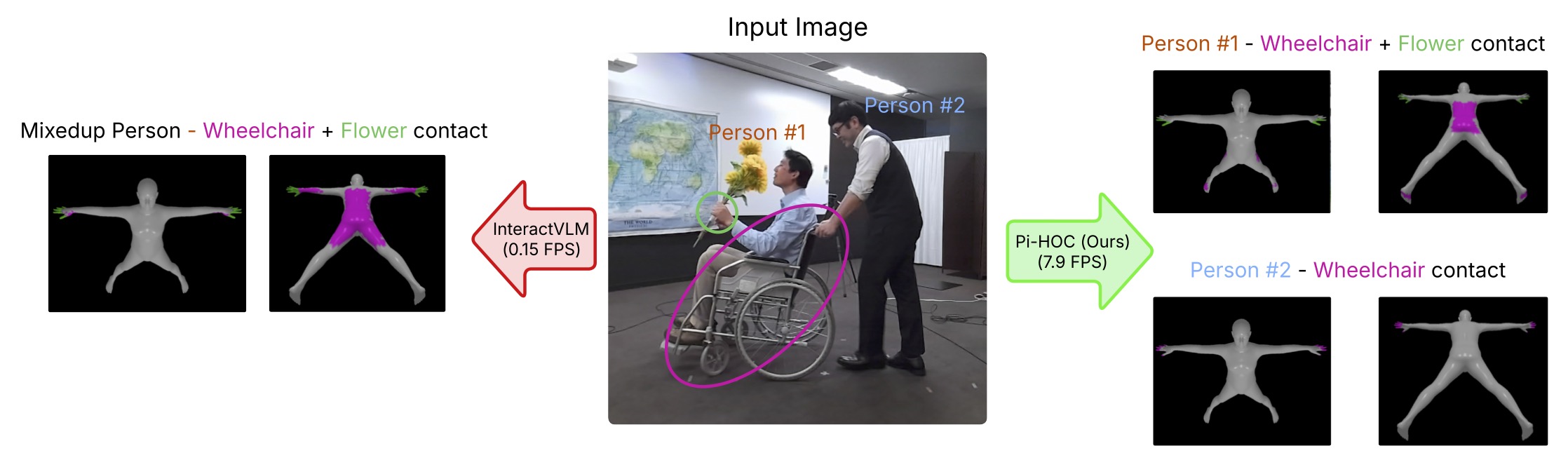}
  \end{center}
  \caption{\textbf{Instance-level contact (our Pi-HOC) vs. Category-level contact (InteractVLM).} In a scene with concurrent interaction among multiple humans and objects, multiple contacts occur in the circled regions. \textbf{Left:} InteractVLM is unaware of instances of the same categories, and mixed up all concurrent contacts of three categories: human, flower, and wheelchair. \textbf{Right:} Pi-HOC correctly parses contacts over multiple human-object instance pairs, assigning each object contact to the correct person. Pi-HOC is also substantially faster (7.9FPS) than InteractVLM (0.15FPS). }
  \label{fig:intro_fig}
\end{figure*}

Human interactions in real-world scenarios are rarely one-to-one. For example, two people may interact with the same wheelchair, one seated while another pushes (Fig.~\ref{fig:intro_fig}). Capturing such interactions is important for human--scene reconstruction~\cite{liu2026joint,ym2025physic,hassan2019resolving,coin2022zhao} and human--object interaction understanding~\cite{nam2024contho,Cseke_2025_CVPR,xie2022chore}, with applications in augmented reality, robotics, and surveillance.

Recent work has progressed from binary scene-level contact classification to dense contact localization in SMPL~\cite{SMPL:2015}. InteractVLM~\cite{dwivedi2025interactvlm} further predicts semantic contacts for a human interacting with multiple objects. Given an image and object name, it uses a large VLM to guide an SAM-based decoder, and lifts the predicted contact to 3D. However, it is limited to single-human scenes. It cannot naturally disambiguate multiple instances of the same object category, and it is computationally expensive.

We address these limitations with Pi-HOC, a single-pass instance-aware semantic contact model for \emph{all} human--object pairs in a scene. Our object-centric approach detects all humans and objects, enumerates all pairs, and represents each pair with a dedicated human--object (HO) token. These tokens are jointly refined with image features using an InteractionFormer, and a contact decoder predicts dense per-vertex contact on a canonical human mesh for every pair. This design enables efficient instance-level reasoning without large VLMs.

We evaluated Pi-HOC on two complementary datasets. MMHOI~\cite{kogashi2025mmhoi} tests contact prediction in multi-person, multi-object scenes, where instance disambiguation is critical. DAMON~\cite{tripathi2023deco} contains in-the-wild images with a single human but multiple candidate objects, emphasizing fine-grained semantic contact. In both settings, Pi-HOC improves contact accuracy and localization while achieving substantially higher throughput than state-of-the-art methods~\cite {dwivedi2025interactvlm}, making it practical for real-time use.

We further demonstrate two downstream applications. First, we improve SAM3D~\cite{sam3dteam2025sam3d3dfyimages,yang2026sam3dbody} image-to-mesh reconstruction by adding contact-based constraints at test time, correcting fine-grained errors such as missing hand--object contacts. Second, we show referential contact prediction: given a free-form natural-language query that specifies a human--object interaction, we predict contact only for the referenced pair in a zero-shot manner, without additional training.

\paragraph{Contributions.}
Our key contributions are as follows.
\begin{itemize}
  \item \textbf{Multi-human, multi-object semantic contact.} We tackle the task of predicting dense, per-vertex semantic contact for every human--object pair in a scene.
  \item \textbf{Single-pass, instance-aware model.} We propose Pi-HOC, which forms HO pairs and refines them with an InteractionFormer and a SAM-based contact decoder, enabling contact prediction for all human--object pairs in one forward pass.
  \item \textbf{State-of-the-art contact prediction and real-time efficiency.} Experiments on MMHOI and DAMON show significant gains in semantic contact accuracy and localization (over 10\% F1), with inference speeds around 20$\times$ faster than the previous state of the art~\cite{dwivedi2025interactvlm}.
  \item \textbf{Applications beyond contact prediction.} We demonstrate referential contact prediction and improvements to SAM-3D reconstruction using our predicted contact.
\end{itemize}

\noindent We make our code publicly available at \url{https://pi-hoc.github.io/}.

\section{Related work}

\subsection{Human--object interaction (HOI) detection}
HOI detection aims to detect and recognize interactions as triplets $\langle$human, object, verb$\rangle$~\cite{li2024bilateral,zhang2022upt,cao2023unihoi,kim2021hotr,ting2023hoi,kim2025lain}.
A common pipeline first detects humans and objects using an off-the-shelf detector (e.g., DETR~\cite{carion2020detr}), enumerates candidate human--object pairs, and then classifies each pair with an interaction verb.
While HOI detection provides interaction semantics for human--object pairs in a scene, it offers limited supervision about \emph{where} contact occurs on the human body surface.
As a result, HOI predictions alone are often insufficient for downstream tasks such as 3D reconstruction and fine-grained scene understanding, which require localized, surface-level contact cues~\cite{nam2024contho,liu2024easyhoi}.
Pi-HOC addresses this gap by going beyond coarse interaction recognition and localizing fine-grained contact regions for each human--object pair.

\subsection{Binary contact classification}
Early contact estimation methods~\cite{chen2023hot,mueller2021selfcontact,fieraru2021complex,hassan2019resolving,hassan2021posa,shimada2022hulc,huang2022rich,tripathi2023deco} focus on binary contact prediction (contact vs. non-contact), rather than contact semantics.
Representations range from 2D image-space masks to dense 3D per-vertex labels on the body mesh. HOT~\cite{chen2023hot} provides crowd-sourced 2D contact annotations and predicts image-aligned contact heatmaps. Although easy to annotate, these labels do not establish direct correspondence on the 3D human surface. Later work predicts binary contact directly on SMPL vertices~\cite{SMPL:2015}. POSA~\cite{hassan2021posa} models contact conditioned on body pose using a conditional generative model(cVAE), while BSTRO~\cite{huang2022rich} formulates contact as structured per-vertex prediction using image features.
RICH~\cite{huang2022rich} supports this line of work with high-quality human--scene reconstructions indoors and outdoors, and DAMON/DECO~\cite{tripathi2023deco} extend supervision and learning to in-the-wild human--object settings.
Overall, these methods estimate dense binary contact but do not explicitly identify contacted object categories or instances.
Our method addresses this gap with object-centric, vertex-level contact prediction tied to specific object instances.

\subsection{Semantic contact estimation}
\label{sec:related-semantic}

In this paper, \emph{semantic contact} refers to predicting (i) a binary per-vertex contact label on the human body surface and (ii) the semantic category of the contacted object.
CONTHO~\cite{nam2024contho} incorporates contact cues when jointly reasoning about humans and objects for 3D reconstruction, predicting contact on both human and object surfaces.
LEMON~\cite{yang2024lemon} exploits correlations between interaction counterparts to jointly predict human contact, object affordance, and 3D human--object spatial relations.
Such approaches typically require paired supervision between human contact and object affordance during training and often assume access to an object point cloud or mesh, in addition to the image, at inference time.
InteractVLM~\cite{dwivedi2025interactvlm} (closest to ours) uses a vision--language model (VLM) with a CLIP backbone, a 13B-parameter LLM, and an object-conditioned prompt, decoding a dedicated \texttt{<HCONTACT>} token to predict human contact. However, object-conditioned prompting and the lack of unambiguous human/object references limit disambiguation between multiple humans and object instances of the same category. It is also computationally expensive due to its large VLM and requires one forward pass for each human--object pair. In contrast, our lightweight model predicts contact for multiple human--object pairs from a single image in one forward pass.



\section{Method}

\begin{figure*}[t]
  \centering
    \includegraphics[width=\textwidth]{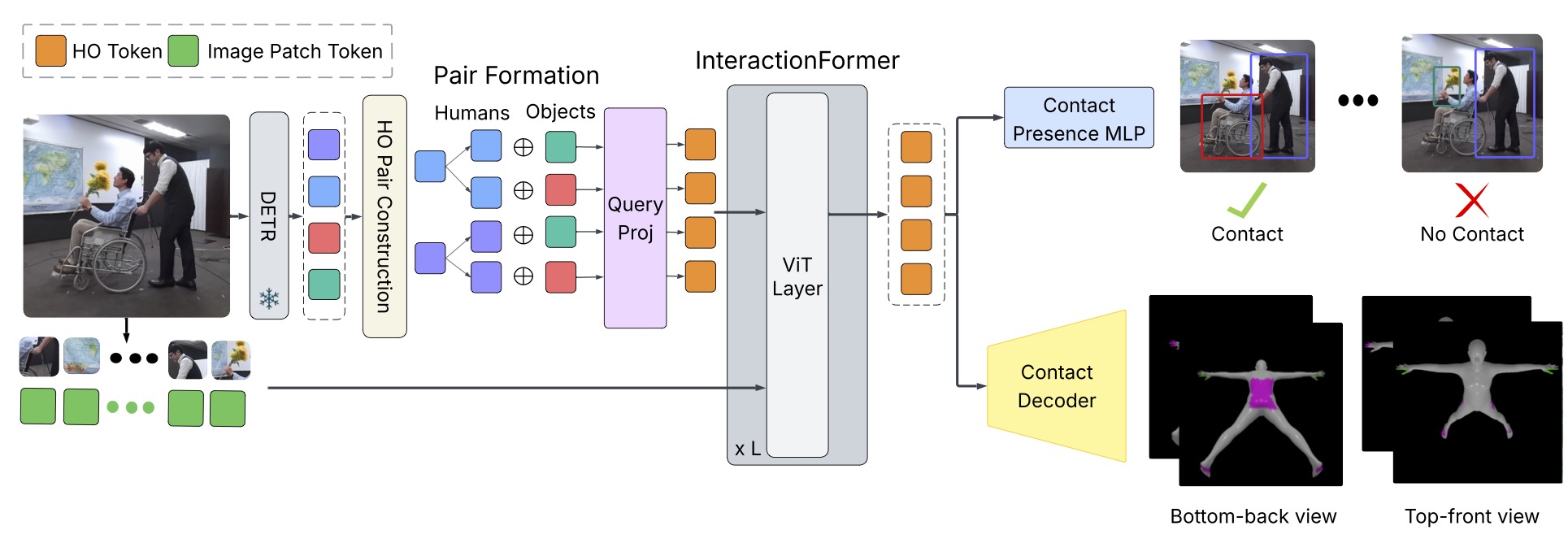}
  \caption{\textbf{Pi-HOC Architecture.}
    Given an input image $I$, a frozen DETR detector localizes human and object instances. A human--object pair constructor enumerates candidate pairs and encodes them as HO tokens (orange). InteractionFormer jointly refines HO tokens and image patch tokens (green) over $L$ blocks. A contact-presence MLP first filters non-contact pairs, and a contact decoder then predicts a per-vertex contact mask $\mathbf{m}_{h,o}\in\{0,1\}^{N_v}$ for each retained pair, yielding contact instances $\tau(h,o)$ ($N_v$: number of human-mesh vertices).}
  \label{fig:arch_det}
\end{figure*}

Given an RGB image $I$, our model predicts fine-grained human--object contact on the human mesh. For each person, it identifies the contacted object instances and the corresponding body-submesh contact regions.

\subsection{Problem formulation}
We decompose the task into two sub-problems: (1) resolving instance-level contact among multiple human-object pairs and (2) localizing fine-grained contact regions on the human body. We define the key terms below.

\begin{itemize}
  \item \textbf{Human and object instances.} Let $\mathcal{H}$ and $\mathcal{O}$ denote the sets of human and object instances in input image $I$. For $h\in\mathcal{H}$ and $o\in\mathcal{O}$, their 2D boxes are $\mathbf{b}_h\in\mathbb{R}^4$ and $\mathbf{b}_o\in\mathbb{R}^4$, with coordinates $(x_\text{min},y_\text{min},x_\text{max},y_\text{max})$. Each object instance has a category label $c_o \in \mathcal{C}$, where $\mathcal{C}$ is the set of object categories.

  \item \textbf{Human Body Mesh.} Each human instance $h$ is represented as a SMPL mesh~\cite{SMPL:2015} with $N_v=6890$ vertices. The vertex coordinates $\mathbf{V}_h\in\mathbb{R}^{N_v\times 3}$ are in a canonical ``star" pose.

  \item \textbf{Contact Mask.} For each human--object pair $(h,o)$, our method predicts a binary per-vertex contact mask $\mathbf{m}_{h,o}\in\{0,1\}^{N_v}$, where $\mathbf{m}_{h,o}(j)=1$ indicates that the human mesh vertex $j$ is in contact with the object $o$, and $\mathbf{m}_{h,o}(j)=0$ otherwise.

  \item \textbf{Contact Instances.} A person $h$ and an object $o$ in contact are defined as a \emph{contact instance}
    \begin{equation}
      \label{eq:contact-instance}
      \tau(h,o)=\bigl(\mathbf{b}_h,\,\mathbf{b}_o,\,c_o,\,\mathbf{m}_{h,o}\bigr).
    \end{equation}
\end{itemize}

\paragraph{\textbf{Formulated problem.}} Based on the terminology above, our goal is to detect all contact instances $\{\tau(h,o)\mid h\in\mathcal{H},\ o\in\mathcal{O}\}$ from an input image $I$.

\paragraph{\textbf{Model overview.}}

Figure~\ref{fig:arch_det} summarizes the Pi-HOC pipeline.
Given an input image $I$, we first apply a frozen DETR detector~\cite{carion2020detr} to obtain human/object instances ($\mathcal{H},\mathcal{O}$) and their boxes ($\mathbf{b}_h,\mathbf{b}_o$). We then enumerate candidate pairs $(h,o)$ and encode each as a human-object (HO) token(Sec.~\ref{sec:ho-token}). $L$ InteractionFormer blocks (Sec.~\ref{sec:ho-refine}) jointly refine HO tokens, accompanied by image patch tokens.
Finally, the contact decoder (Sect.~\ref{sec:decoder}) predicts a binary per-vertex contact mask $\mathbf{m}_{h,o}\in\{0,1\}^{N_v}$ for each contact instance $\tau(h,o)$ in Eq.~\eqref{eq:contact-instance}.

\subsection{HO Token Construction}
\label{sec:ho-token}

\paragraph{Post-processing DETR detections.}
We start from DETR predictions and apply standard post-processing (confidence thresholding and non-maximum suppression) to obtain a filtered set of detections
\begin{equation}
  \mathcal{D}=\bigl\{(\mathbf{b}_k,\, s_k,\, c_k,\, \mathbf{q}_k)\bigr\}_{k=1}^{N},
\end{equation}
where $\mathbf{b}_k\in\mathbb{R}^4$ is the 2D bounding box, $s_k\in[0,1]$ is the confidence score, $c_k\in\mathcal{C}$ is the category label and $\mathbf{q}_k\in\mathbb{R}^{D_{\text{DETR}}}$ is the DETR query feature.
Using class labels, we partition $\mathcal{D}$ into human detections $\mathcal{D}^h$ and object detections $\mathcal{D}^o$ (i.e., $\mathcal{D}^h\cap\mathcal{D}^o=\emptyset$).

\paragraph{Constructing Human--Object Pairs.}
We pair each human with each object and prune non-overlapping boxes, since such pairs cannot be in contact.
We define
\begin{equation}
  \mathcal{P}=\Bigl\{(h,o)\,\big|\, h\in\mathcal{D}^h,\ o\in\mathcal{D}^o,\ \mathrm{IoU}(\mathbf{b}_h,\mathbf{b}_o)>\gamma\Bigr\},
\end{equation}
where $\gamma\ge 0$ is an IoU threshold (we use $\gamma=0$ to keep overlapping pairs only).

\paragraph{Embedding Human--Object (HO) Tokens.}
To model instance-specific interactions, we explicitly bind each detected human to an object candidate. For each pair $(h,o)\in\mathcal{P}$, we construct an HO token by projecting concatenated DETR query features:
\begin{equation}
  \mathbf{t}_{h,o}=\phi\bigl([\mathbf{q}_h;\mathbf{q}_o]\bigr)\in\mathbb{R}^{D_{\text{IF}}},
\end{equation}
where $[\cdot;\cdot]$ denotes concatenation and $\phi$ is a small feed-forward MLP.
Stacking all pair tokens yields $\mathbf{T}\in\mathbb{R}^{|\mathcal{P}|\times D_{\text{IF}}}$, which is fed to InteractionFormer together with the image patch tokens (Fig.~\ref{fig:arch_det}).

\subsection{Refining HO Tokens}
\label{sec:ho-refine}
\begin{figure}[t]
  \centering
  \includegraphics[width=1.00\linewidth]{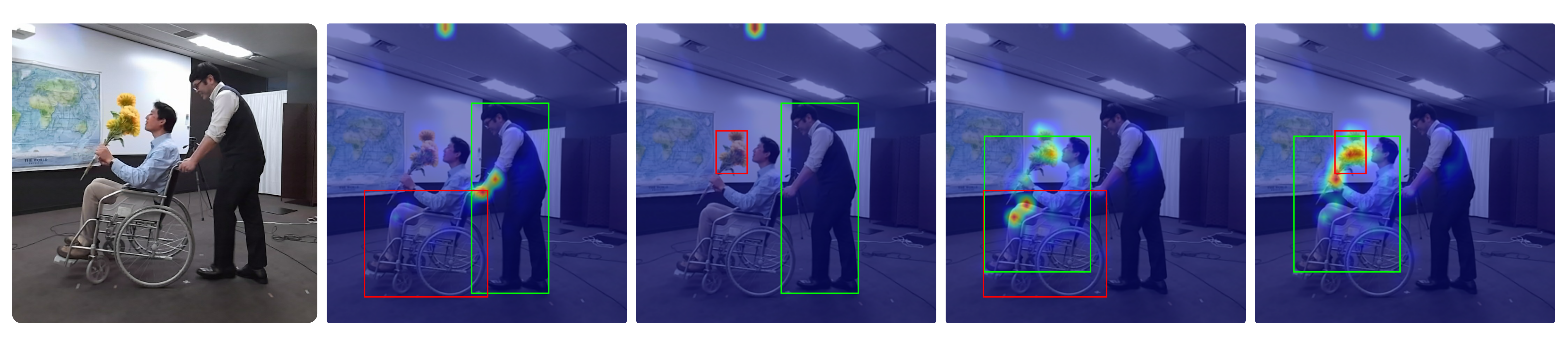}
  \caption{\textbf{InteractionFormer Attention Visualization.} Attention maps show that each HO token, $\mathbf{t}_{h,o}$, attends to the image regions corresponding to its associated human--object pair and also surrounding regions for additional context. The green bounding box indicates the human instance. The red bounding box indicates the object instance in the corresponding HO pair.}
  \label{fig:attn_viz}
\end{figure}

We use a transformer encoder, which we call InteractionFormer, to jointly refine the HO tokens and the image patch tokens via multi-head self-attention layers.
Concretely, each InteractionFormer \emph{block} is a standard Transformer encoder layer (self-attention + MLP); we apply $L$ blocks sequentially.
We initialize these blocks from pretrained DINOv2~\cite{oquab2023dinov2} weights and fine-tune them for contact estimation.
The intuition is that, although the initial HO token $\mathbf{t}_{h,o}$ only binds a particular human $h$ and an object $o$ through detector features, repeated self-attention updates allow $\mathbf{t}_{h,o}$ to attend to the relevant image evidence around the pair and encode (implicitly) the interaction type, involved body parts, and approximate contact location, making the HO tokens interaction-aware.
Figure~\ref{fig:attn_viz} visualizes this behavior in the last InteractionFormer block using attention maps, showing that each HO token attends to the image regions corresponding to its associated human--object pair.

We denote by $N_p$ the number of image patch tokens and by $D_{\text{IF}}$ the dimension of the InteractionFormer feature (in our implementation, $D_{\text{IF}}=1024$).
Let $\mathbf{X}\in\mathbb{R}^{N_p\times D_{\text{IF}}}$ denote the image patch tokens extracted from $I$.
We form the input token sequence $\mathbf{Z}^{(\ell)}$ in layer $\ell=0$ as
\begin{equation}
  \mathbf{Z}^{(0)}=\bigl[\mathbf{T}^{(0)};\,\mathbf{X}\bigr],\qquad \mathbf{T}^{(0)}\equiv \mathbf{T},
\end{equation}
where $[\cdot;\cdot]$ denotes sequence concatenation. For $\ell=0,\dots,L-1$, the token sequence is refined by the $\ell$-th InteractionFormer block, denoted by $\mathrm{IF}_{\ell}(\cdot)$.
Writing the sequence explicitly as
$\mathbf{Z}^{(\ell)}=[\mathbf{T}^{(\ell)};\,\mathbf{X}^{(\ell)}]$, the update can be expressed as
\begin{equation}
  \label{eq:if-update}
  \bigl[\mathbf{T}^{(\ell+1)};\,\mathbf{X}^{(\ell+1)}\bigr]
  =\mathrm{IF}_{\ell}\Bigl(\bigl[\mathbf{T}^{(\ell)};\,\mathbf{X}^{(\ell)}\bigr])
\end{equation}
where $\mathbf{T}^{(\ell+1)}\in\mathbb{R}^{|\mathcal{P}|\times D_{\text{IF}}}$ corresponds to the first $|\mathcal{P}|$ output tokens (i.e., the HO-token subsequence).
After $L$ layers, we obtain refined HO tokens $\mathbf{T}^{(L)}$ that are interaction-aware through attention to relevant image context.

\subsection{Contact Decoder}
\label{sec:decoder}

Given the refined HO tokens (Sec.~\ref{sec:ho-refine}), we predict a binary per-vertex contact mask $\mathbf{m}_{h,o}\in\{0,1\}^{N_v}$ for each candidate pair $(h,o)\in\mathcal{P}$. Directly decoding all 6,890 SMPL vertices is computationally expensive.
Following InteractVLM~\cite{dwivedi2025interactvlm}, we decode HO tokens into multi-view 2D contact maps, where each map represents pixel-wise contact likelihood on a 2D projection of the SMPL mesh. We then lift these maps back to the 3D SMPL surface. Additional contact-decoder details are provided in the supplementary material.

\paragraph{Classifying Contact Presence.}
In practice, most candidate pairs $(h,o)\in\mathcal{P}$ are not in physical contact (e.g., overlapping background objects), so running the SAM decoder for every pair is inefficient.
We therefore attach a lightweight contact-presence head that predicts whether a pair is in contact.
Specifically, the head is an MLP classifier $f_{\text{cp}}(\cdot)$ on the refined HO token and produces a binary contact probability:
\begin{equation}
  p_{h,o}=f_{\text{cp}}\bigl(\mathbf{t}^{(L)}_{h,o}\bigr)\in[0,1].
\end{equation}
In inference, we keep only pairs that are prone to contact with $p_{h,o}>\delta$ and run the SAM-based view decoder only for them, improving efficiency and focusing computation on likely contacts.

\subsection{Training Objectives}
\label{sec:training-objectives}

\paragraph{Overview of Loss Terms.}
We optimize four loss components: (i) a view-wise 2D contact loss $\mathcal{L}_{\text{2D}}$, (ii) a vertex-level 3D contact loss $\mathcal{L}_{\text{3D}}$, (iii) a contact-presence loss $\mathcal{L}_{\text{cp}}$, and (iv) Following DECO~\cite{tripathi2023deco}, optional auxiliary scene/body-part segmentation losses $\mathcal{L}^{s}_{\text{2D}}$ and $\mathcal{L}^{p}_{\text{2D}}$ when segmentation GT is available.
For completeness, $\mathcal{L}_{\text{2D}}$ combines focal and Dice terms, while $\mathcal{L}_{\text{3D}}$ combines a vertex focal term and a $\ell_1$ sparsity term.
We defer full pixel-/vertex-level formulations and implementation details of these losses to the supplementary material.

\paragraph{Contact Presence Supervision.}
We additionally supervise the contact presence classifier (Sec.~\ref{sec:decoder}) with a binary cross-entropy loss:
\begin{equation}
\mathcal{L}_{\text{cp}}=\sum_{(h,o)\in\mathcal{P}} \mathrm{BCE}\bigl(p_{h,o},\, y_{h,o}^{\text{cp}}\bigr),
\end{equation}
where $y_{h,o}^{\text{cp}}\in\{0,1\}$ indicates whether the pair $(h,o)$ is in contact.

The total training loss is
\begin{equation}
\mathcal{L}=\mathcal{L}_{\text{2D}}+\mathcal{L}_{\text{3D}}+\lambda_{\text{cp}}\,\mathcal{L}_{\text{cp}}+\mathbb{I}_{\mathrm{GT}}\Bigl(\lambda_s\,\mathcal{L}^{s}_{\text{2D}}+\lambda_p\,\mathcal{L}^{p}_{\text{2D}}\Bigr),
\end{equation}
where $\mathbb{I}_{\mathrm{GT}}\in\{0,1\}$ indicates whether segmentation GT supervision is present and $\lambda_s,\lambda_p$ are the corresponding loss weights.
\section{Experiments} \label{sec:exps}

\begin{figure*}[t]
  \begin{center}
    \includegraphics[width=\textwidth]{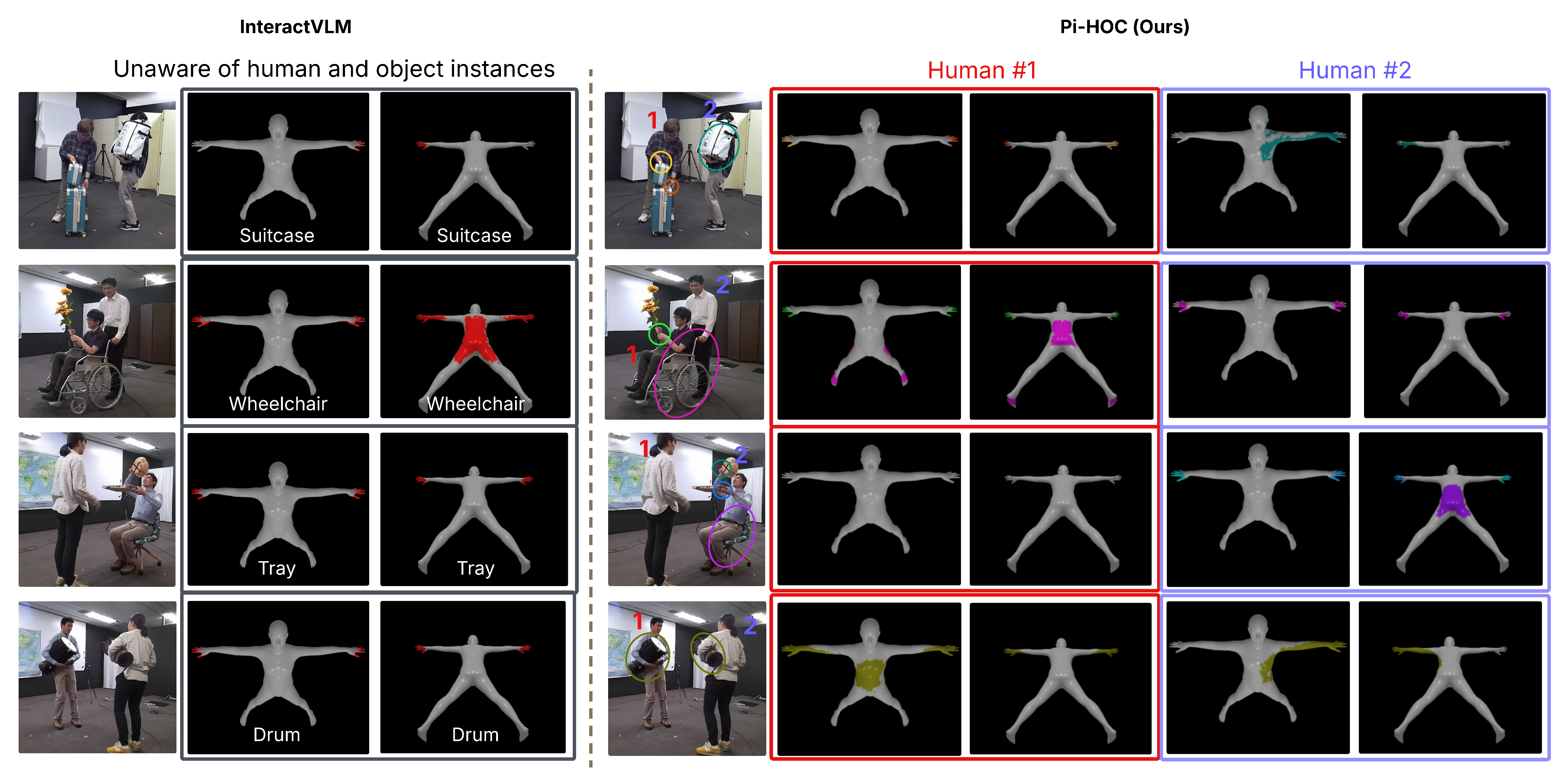}
  \end{center}
  \caption{\textbf{Qualitative Comparison.} Left: InteractVLM, which is object-category conditioned, predicts contacts (red) but fails to disambiguate multiple people or multiple instances of objects of the same category. Right: Pi-HOC correctly distinguishes individuals and predicts contacts for each human-object pair. Human instance id is shown next to each person, and contacted objects are circled in distinct colors.}
  \label{fig:qualitative_interactvlm_vs_pi-hoc}
\end{figure*}

\begin{table}[t]
  \centering
  \small
  \caption{\textbf{MMHOI Semantic Contact Results}.}
  \label{tab:mmhoi_results}
  \setlength{\tabcolsep}{3pt}
  \resizebox{\columnwidth}{!}{%
    \begin{tabular}{lcccc}
      \toprule
      Method & F1 $\uparrow$ & Prec $\uparrow$ & Recall $\uparrow$ & Geo. dist $\downarrow$ \\
      \midrule
      InteractVLM-Crop$_{\text{Pretrained}}$ & 43.68 & 39.23 & 62.13 & 0.2468 \\
      InteractVLM$_{\text{Trained}}$  & 52.02 & 48.89 & 69.25 & 0.084 \\
      InteractVLM-Crop$_{\text{Trained}}$  & 54.86 & 49.91 & 69.40 & 0.073 \\
      Pi-HOC (Ours)                   & \textbf{61.09} & \textbf{58.76} & \textbf{71.98} & \textbf{0.0633} \\
      \bottomrule
    \end{tabular}%
  }
\end{table}

\begin{table}[t]
  \centering
  \small
  \caption{\textbf{DAMON Semantic Contact Results}.}
  \label{tab:damon_results}
  \setlength{\tabcolsep}{3pt}
  \resizebox{\columnwidth}{!}{%
    \begin{tabular}{lcccc}
      \toprule
      & F1 $\uparrow$ & Prec $\uparrow$ & Recall $\uparrow$ & Geo. dist $\downarrow$ \\
      \midrule
      Semantic DECO~\cite{tripathi2023deco}
      & 32.72 & 25.91 & \textbf{71.51} & 0.59 \\

      InteractVLM-w/o-body-parts\footref{fn:interactvlmfootnote}~\cite{dwivedi2025interactvlm}
      & \underline{59.48} & \underline{59.36} & 68.06 & \underline{0.095} \\

      Pi-HOC (Ours)
      & \textbf{63.23} & \textbf{64.49} & \underline{69.57} & \textbf{0.092} \\
      \bottomrule
    \end{tabular}%
  }
\end{table}

\begin{table}[t]
  \centering
  \small
  \caption{\textbf{Inference Speed Comparison.} FPS as a function of the number of human--object pairs per image.}
  \label{tab:inference_speed}
  \setlength{\tabcolsep}{4pt}
  \resizebox{\columnwidth}{!}{%
    \begin{tabular}{lccccccc}
      \toprule
      \# Pairs & 1 & 2 & 3 & 5 & 7 & 9 & 11 \\
      \midrule
      InteractVLM~\cite{dwivedi2025interactvlm} & 0.30 & 0.15 & 0.096 & 0.057 & 0.041 & 0.032 & 0.026 \\
      Pi-HOC (Ours) & \textbf{8.5} & \textbf{7.9} & \textbf{6.5} & \textbf{5.1} & \textbf{4.0} & \textbf{3.0} & \textbf{2.3} \\
      \bottomrule
    \end{tabular}%
  }
\end{table}

\begin{table*}[t]
  \centering
  \caption{\textbf{Ablation Studies.} (a) InteractionFormer initialization. (b) InteractionFormer layer depth. (c) Contact decoder design. (d) Auxiliary losses.}
  \label{tab:ablations}
  \small
  \setlength{\tabcolsep}{3pt}
  \renewcommand{\arraystretch}{1.15}

  \begin{subtable}[t]{0.49\textwidth}
    \centering
    \caption{\textbf{InteractionFormer Init.}}
    \label{tab:ablation_init}
    \begin{tabular}{lccc}
      \toprule
      IF Init. & F1 $\uparrow$ & Prec $\uparrow$ & Recall $\uparrow$ \\
      \midrule
      Random Init.   & 56.44 & 51.95 & 71.32 \\
      CLIP-L   & \underline{58.8} & \underline{55.26} & \underline{71.38} \\
      DINOV2-L & \textbf{61.09} & \textbf{58.76} & \textbf{71.98} \\
      \bottomrule
    \end{tabular}
  \end{subtable}
  \hfill
  \begin{subtable}[t]{0.49\textwidth}
    \centering
    \caption{\textbf{InteractionFormer Layer Depth}}
    \label{tab:ablation_if_depth}
    \begin{tabular}{lccc}
      \toprule
      IF depth & Prec $\uparrow$ & Recall $\uparrow$ & Geo. dist $\downarrow$ \\
      \midrule
      0 / 24 (none)     & 53.27 & 71.41 & 7.02 \\
      first 6 / 24      & \underline{54.33} & \textbf{72.27} & \underline{6.98} \\
      24 / 24 (full)    & \textbf{58.76} & \underline{71.98} & \textbf{6.33} \\
      \bottomrule
    \end{tabular}
  \end{subtable}

  \vspace{0.8em}

  \begin{subtable}[t]{0.49\textwidth}
    \centering
    \caption{\textbf{Contact Decoder Design}}
    \label{tab:ablation_decoder}
    \begin{tabular}{lccc}
      \toprule
      Decoder & Prec $\uparrow$ & Recall $\uparrow$ & Geo. dist $\downarrow$ \\
      \midrule
      MLP         & 57.05 & 58.35 & 12.54 \\
      SAM Decoder & \textbf{58.76} & \textbf{71.98} & \textbf{6.33} \\
      \bottomrule
    \end{tabular}
  \end{subtable}
  \hfill
  \begin{subtable}[t]{0.49\textwidth}
    \centering
    \caption{\textbf{Auxiliary Losses}}
    \label{tab:ablation_aux}
    \begin{tabular}{lcccc}
      \toprule
      $\mathcal{L}_{2D}^{s}$ & $\mathcal{L}_{2D}^{p}$ & F1 $\uparrow$ & Geo. dist $\downarrow$ \\
      \midrule
      $\times$  & $\times$  & 62.06 & 12.46 \\
      $\checkmark$ & $\times$  & 62.35 & 11.79 \\
      $\times$  & $\checkmark$ & 62.66 & 10.49 \\
      $\checkmark$ & $\checkmark$ & \textbf{63.28} & \textbf{9.24} \\
      \bottomrule
    \end{tabular}
  \end{subtable}

\end{table*}

\subsection{Experimental Setup}
\paragraph{Datasets.}
We evaluated on two datasets: MMHOI~\cite{kogashi2025mmhoi} and DAMON~\cite{tripathi2023deco}.
MMHOI contains multiple humans (typically 2--3) that interact with one or more objects.
Since MMHOI does not provide contact annotations, we generated contact annotations by thresholding the distance between the human and object meshes at 2 cm.
We split MMHOI into 18k training images and 4k test images.
DAMON consists of 4k training images and 800 test images, which were manually annotated with semantic contact labels.
It is a standard dataset that is primarily used to evaluate the performance of single-person contacts.
Together, these datasets allow us to evaluate both challenging cases with multiple interacting person-object pairs (MMHOI) and single-person cases with multiple candidate objects (DAMON).

\paragraph{Metrics.}
We report precision, recall, and F1 at a contact threshold of 0.5.
Following ~\cite{huang2022rich,tripathi2023deco,dwivedi2025interactvlm}, we also report geodesic distance on the SMPL surface as a geometry-aware localization error.
For each predicted contact vertex, we compute the shortest geodesic distance to the nearest ground-truth contact vertex. Distant mislocalizations (e.g., contact in the hand vs.\ hips) are penalized more than nearby confusions (e.g., contact in the hand vs.\ forearm), providing a more informative measure of contact localization quality.
Together, these metrics separate contact classification performance from spatial localization accuracy.

\paragraph{Baselines.}
We compare Pi-HOC against Semantic-DECO and InteractVLM-w/o-body-parts~\cite{dwivedi2025interactvlm}\footnote{\label{fn:interactvlmfootnote}The original InteractVLM paper reported metrics that used GT contact body parts at inference, which is an evaluation bug that we verified using official GitHub codes and confirmed with the authors via email. Per their recommendation, we use the version w/o body parts for fair comparison. See the \textbf{supplementary materials} for the original and w/o-body-parts versions that we reported in this section.}. Semantic-DECO extends DECO~\cite{tripathi2023deco} from predicting binary contact to predicting semantic (multi-class) contact. On DAMON, which primarily contains single-person scenes, we directly report the performance of these methods.
For MMHOI, which contains multiple humans and objects, we report numbers on \emph{InteractVLM$_{\text{Trained}}$}, i.e., InteractVLM trained on MMHOI. Because InteractVLM is designed for a single human-object interaction conditioned on an image and an object name, we additionally define a multi-person baseline, \emph{InteractVLM-Crop}. For each human-object pair, we crop the image to the tight region covering the union of the corresponding human and object bounding boxes (using GT boxes) and query InteractVLM with the relevant object name.
We consider two variants of InteractVLM-Crop: \emph{InteractVLM-Crop$_{\text{Pretrained}}$} (published pretrained weights) and \emph{InteractVLM-Crop$_{\text{Trained}}$} (trained on MMHOI with the same cropped inputs).

\subsection{Semantic Human Contact}
\paragraph{Multi-human multi-object scenarios.}
Multi-person, multi-object scenes require \emph{instance-aware} reasoning to (i) associate each person with the correct object(s) and (ii) localize contact on the correct body instance.
Prior methods such as InteractVLM are not designed for this setting and tend to mis-assign contact based only on object category.
Table~\ref{tab:mmhoi_results} shows that Pi-HOC outperforms all multi-person baselines on MMHOI. Relative to \emph{InteractVLM-Crop$_{\text{Trained}}$}, it improves F1 by 11.4\% (54.86 $\rightarrow$ 61.09) and reduces geodesic distance by 13.3\% (0.073 $\rightarrow$ 0.0633; 1.15$\times$ lower).
We attribute this gap partly to the \emph{union-crop} strategy in InteractVLM-Crop: each crop can include extra people or nearby objects, creating ambiguity about the queried interaction and hurting both classification and localization (reflected by geodesic error). See Fig.~\ref{fig:qualitative_interactvlm_vs_pi-hoc} for qualitative examples. For more qualitative results, please check the supplementary file.

\paragraph{Single-human Multi-object Scenarios.}
Single-person scenes with multiple candidate objects are the main focus of DAMON.
As shown in Table~\ref{tab:damon_results}, compared to InteractVLM, Pi-HOC improves F1 by 6.3\% (59.48 to 63.23) and reduces the geodesic distance by 3.2\% (0.095 to 0.092; 1.03$\times$ lower), despite InteractVLM using a 13B-parameter language model and training on the same data.
These gains indicate that Pi-HOC predicts contacts that are more accurate and \emph{spatially} closer to the ground truth.
We attribute this improvement in part to stronger spatial features from the pretrained DINOv2 backbone, which provides a finer localization than the CLIP backbone used by InteractVLM (Table~\ref{tab:ablation_init}), especially when multiple objects are near. Additional object-wise comparisons are provided in the supplementary material.

\subsection{Inference Speed}

Table~\ref{tab:inference_speed} shows that Pi-HOC is consistently more than an order of magnitude faster than InteractVLM across all numbers of human--object pairs. For example, Pi-HOC reaches 8.5 vs.~0.30 FPS for one pair (28.3$\times$ faster) and 2.3 vs.~0.026 FPS for 11 pairs (88.5$\times$ faster).
Pi-HOC also degrades more gracefully as pair count increases (8.5$\rightarrow$2.3 FPS, 3.7$\times$ drop) than InteractVLM (0.30$\rightarrow$0.026 FPS, 11.5$\times$ drop), indicating substantially better scalability in cluttered multi-person scenes.
This trend is consistent with Pi-HOC's batched pair processing, while InteractVLM requires repeated object-conditioned queries.
As a result, Pi-HOC is better suited for latency-critical deployments such as AR/VR, robotics, and safety/service monitoring.

\subsection{Ablation Study}
We isolate the contribution of Pi-HOC components in ablation experiments below. All ablations are performed on MMHOI unless otherwise stated.
\paragraph{InteractionFormer Initialization.}
We compare three InteractionFormer initializations: random, CLIP-L~\cite{radford2021learning}, and DINOv2-L.
As shown in Table~\ref{tab:ablation_init}, DINOv2-L yields the best overall performance. We attribute this gain to the stronger spatial location of the pretrained DINOv2 features, whereas the pretrained CLIP features are optimized primarily for global image--text alignment and object presence~\cite{zhong2022regionclip,wu2023clipself}. In contrast, random initialization lacks a useful structure at the beginning of training, making fine-grained contact localization more difficult and leading to lower precision.
\paragraph{InteractionFormer depth.}
We vary the number of InteractionFormer refinement blocks (including a variant that removes InteractionFormer entirely) and report the results in Table~\ref{tab:ablation_if_depth}.
We observe that recall is relatively stable across depths, suggesting that the coarse interaction cues are based on object categories, which are already captured in the DETR detector features (eg, for person-chair interaction, the contact is generally on the hips).
In contrast, increasing the number of refinement blocks consistently improves precision, indicating that HO tokens benefit from repeatedly attending to localized image evidence to disambiguate whether (and where) contact occurs.
\paragraph{Contact decoder design.}
We ablate the mask decoder head by comparing the SAM-based decoder used in our model with an MLP-based decoder similar to DECO~\cite{tripathi2023deco}.
Table~\ref{tab:ablation_decoder} shows that the SAM-based decoder substantially outperforms the MLP alternative, highlighting the importance of a strong mask decoding head for fine-grained contact localization.
\paragraph{Auxiliary segmentation losses.}
Finally, we ablate the auxiliary 2D supervision terms on DAMON, since the scene and body part annotations are available only in this dataset. In Table~\ref{tab:ablation_aux}, adding either $\mathcal{L}_{2D}^{s}$ (scene) or $\mathcal{L}_{2D}^{p}$ (body-part) improves over the baseline. In particular, $\mathcal{L}_{2D}^{p}$ produces stronger gains, improving both the F1 and the geodesic distance, suggesting better contact localization. Intuitively, the scene segmentation encourages stronger object-aware features, while body-part supervision sharpens localized, part-aligned representations that help disambiguate fine-grained contacts (e.g., left vs. right hand).

\section{Applications}

\begin{figure}[t]
  \centering
  \includegraphics[width=0.9\linewidth]{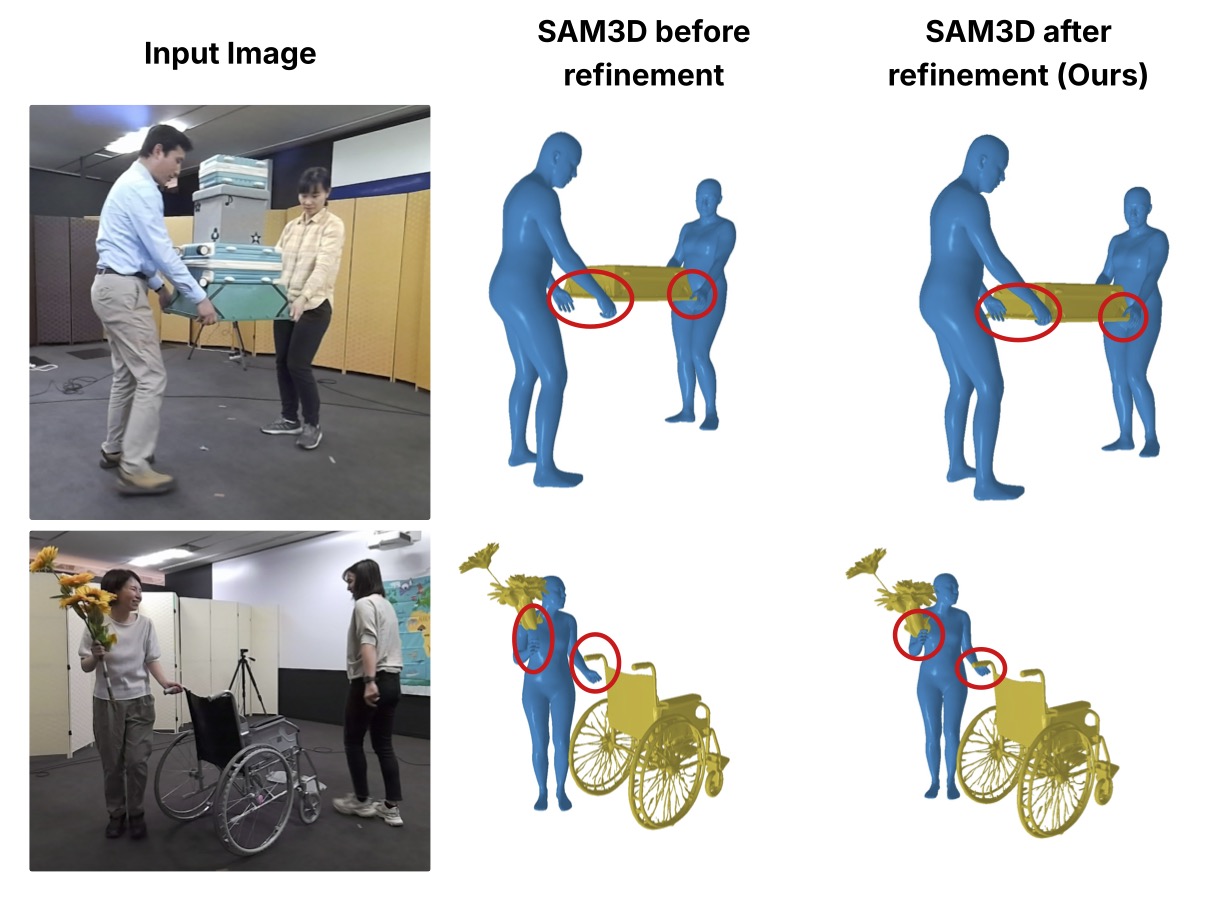}
  \caption{\textbf{SAM-3D test-time refinement with Pi-HOC contact.} Left: input image. Middle: SAM3D Body and Object initialization, which may violate plausible human--object contact (e.g., floating hands). Right: contact-guided refinement with Pi-HOC predictions, yielding more physically consistent interactions.}
  \label{fig:sam3d_refinement}
\end{figure}

\subsection{Improving SAM3D Reconstruction}

SAM3D Object model~\cite{sam3dteam2025sam3d3dfyimages} retrieves object meshes from images and places them in the scene. The SAM3D Body model~\cite{yang2026sam3dbody} retrieves human meshes similarly. Although reconstructions are visually strong, they often violate contact (e.g., floating hands). We therefore propose a test-time refinement that enforces plausible human--object contact.

We initialize body pose and location from SAM3D Body, and object pose and mesh from SAM3D Object. We apply Pi-HOC to detect contact for every human--object pair in the scene. We transfer contacts from the SMPL mesh to the MHR mesh produced by SAM3D. Given the strong SAM3D initialization, we also derive object-side contact by mapping each human contact vertex to its nearest point on the object mesh.

Following InteractVLM~\cite{dwivedi2025interactvlm}, we optimize rotation $\mathbf{R}_h \in SO(3)$, translation $\mathbf{t}_h \in \mathbb{R}^3$, and scale $s_h \in \mathbb{R}^+$ for each human mesh $h$ to better satisfy contacts. At each iteration, we render the current human mesh using PyTorch3D and optimize a weighted objective that combines three cues: mask overlap (IoU), mask-center alignment, and contact consistency between predicted human and object contact regions. We jointly optimize all human--object pairs. The complete objective and equation-level details are provided in the supplementary material.

We compare the original and refined results in Fig.~\ref{fig:sam3d_refinement} (More visualizations in the supplementary material). Before refinement, the reconstructed human and the object are spatially close but do not make physical contact. Incorporating our predicted contact during test-time optimization yields reconstructions with accurate, physically consistent contact.

\subsection{Referential Contact Estimation}

\begin{figure}[t]
  \centering
  \includegraphics[width=1.00\linewidth]{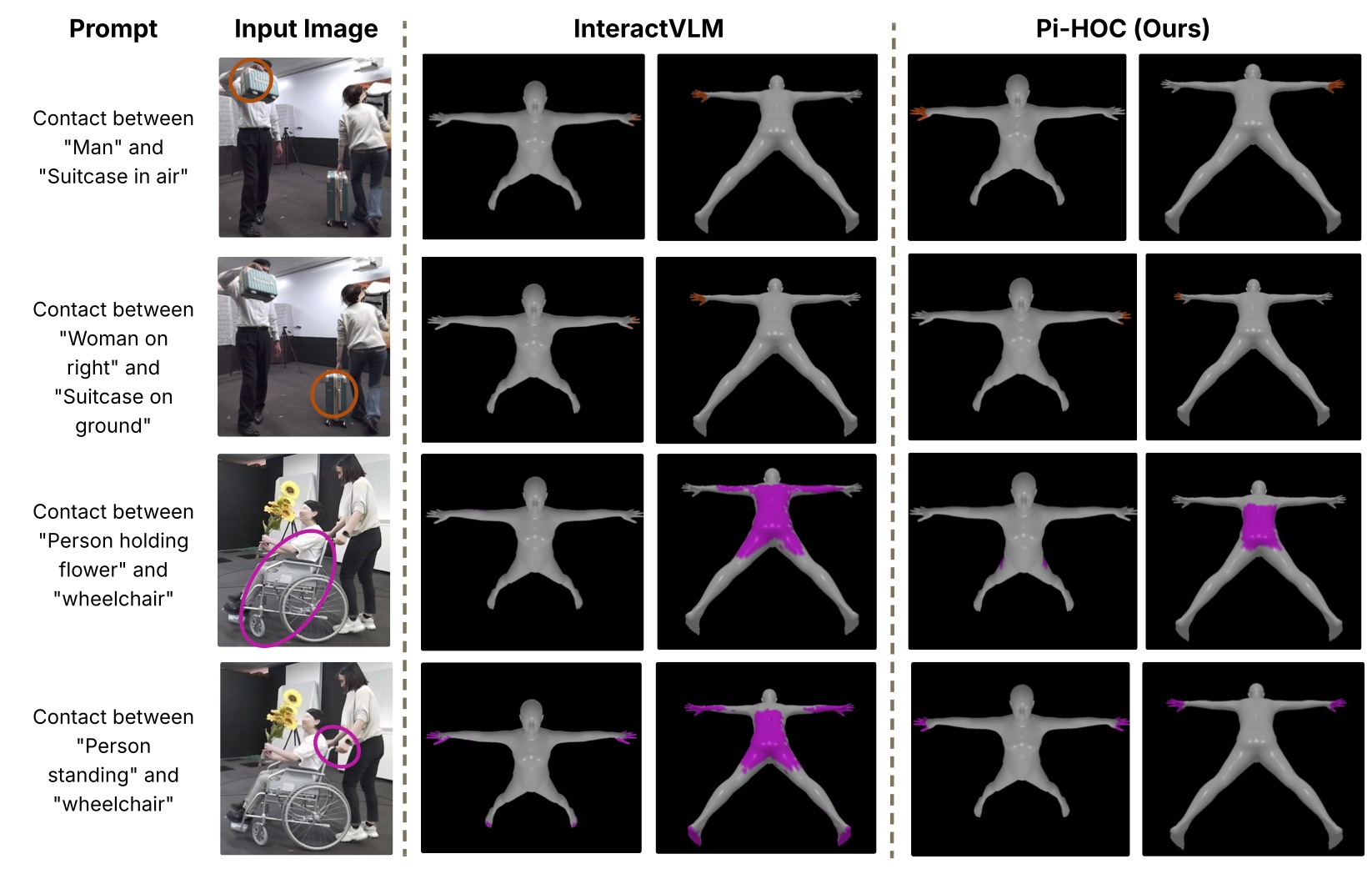}
  \caption{\textbf{Referential contact estimation in multi-instance scenes.} We compare InteractVLM and Pi-HOC under language queries that refer to specific interaction instances. \textbf{Rows 1--2:} queries target different suitcase instances; InteractVLM predicts nearly identical contact and fails to disambiguate the referred suitcase, while Pi-HOC correctly shifts contact from the right hand (row 1) to the left hand (row 2). \textbf{Rows 3--4:} queries target different people interacting with the same object (wheelchair); InteractVLM mixes contacts across both people, whereas Pi-HOC isolates the correct human instance (sitting vs. standing) and predicts pair-specific contact. Colored circles highlight contact regions for the queried interaction.}
  \label{fig:referential_contact}
\end{figure}
Given an image and a natural-language description of a human--object interaction (e.g., ``person sitting on the chair''), the goal \textbf{referential contact estimation} is to predict contact only for the referred human-object pair, rather than for all pairs in the scene. This setting is practically important when downstream applications care about a specific interaction. To our knowledge, no existing dataset provides annotations for this task. 
Owing to our object-centric design, Pi-HOC naturally supports this setting in a zero-shot manner. Given a language query, GroundingDINO~\cite{liu2023grounding} localizes the relevant human and object. We then match these boxes to Pi-HOC detections using maximum IoU, select the corresponding HO pair, and predict contact only for that pair. Qualitative results in Fig.~\ref{fig:referential_contact} show accurate query-conditioned contact prediction for the human-object interaction referred to compared to InteractVLM\cite{dwivedi2025interactvlm}.
\section{Conclusion}

We presented \textbf{Pi-HOC}, a single-pass, instance-aware framework for dense 3D semantic contact prediction in multi-human, multi-object scenes. Motivated by the many-to-many nature of real-world human--object interactions, we address key limitations of prior methods, including single-human assumptions, weak instance awareness, dependence on paired object geometry, reliance on large VLMs, and poor scalability as human--object pairs increase. Pi-HOC adopts an object-centric formulation that explicitly enumerates all human--object (HO) pairs and encodes each pair with a dedicated HO token, enabling efficient and scalable instance-level reasoning. Beyond contact prediction, we show downstream applications: contact-aware test-time optimization to improve SAM-3D image-to-mesh reconstruction and referential contact prediction from language queries without additional training. These results highlight instance-level contact reasoning as a useful structural prior for human--scene understanding. In general, Pi-HOC provides an efficient and scalable solution for predicting dense semantic contacts in complex scenes. We hope this work inspire further research on structured multi-entity interaction modeling for physically grounded scene understanding in augmented reality, robotics, and beyond.

{
  \small
  \bibliographystyle{ieeenat_fullname}
  \bibliography{main}
}

\clearpage
\twocolumn[{%
    \begin{center}
      {\LARGE \textbf{Appendix}}
    \end{center}
    \vspace{0.6em}
}]

This supplementary document is organized as follows. In Sec.~\ref{sec:supp-interactvlm-eval}, we present the corrected InteractVLM evaluation protocol. In Sec.~\ref{sec:supp-method-details}, we detail our method, including implementation details (Sec.~\ref{sec:supp-implementation-details}), the contact decoder (Sec.~\ref{sec:supp-decoder}), training objectives (Sec.~\ref{sec:supp-training-objectives}), and SAM3D test-time refinement (Sec.~\ref{sec:supp-sam3d-refinement}). We then report object-wise semantic contact metrics in Sec.~\ref{sec:supp-objectwise-metrics}, followed by qualitative results in Sec.~\ref{sec:supp-qualitative} and failure cases in Sec.~\ref{sec:supp-failure-cases}. Finally, in Sec.~\ref{sec:supp-attn}, we visualize InteractionFormer attention maps to provide intuition about which regions the model relies on for contact prediction.

\renewcommand{\thesection}{S.\arabic{section}}
\renewcommand{\thesubsection}{S.\arabic{section}.\arabic{subsection}}

\section{InteractVLM Evaluation Protocol}
\label{sec:supp-interactvlm-eval}

\begin{table}[t]
  \centering
  \small
  \caption{\textbf{DAMON Semantic Contact Results}.}
  \label{tab:supp_damon_results}
  \setlength{\tabcolsep}{3pt}
  \resizebox{\columnwidth}{!}{%
    \begin{tabular}{lcccc}
      \toprule
      & F1 $\uparrow$ & Prec $\uparrow$ & Recall $\uparrow$ & Geo. dist $\downarrow$ \\
      \midrule
      InteractVLM (+Part Sup., original eval)
      & 71.2 & 72.91 & 76.29 & 0.075 \\

      InteractVLM (+Part Sup., corrected eval)
      &  57.9 & 59.15 & 64.24 & 0.213 \\

      InteractVLM-w/o-body-parts
      & 59.48 & 59.36 & 68.06 & 0.095 \\

      Pi-HOC (Ours)
      & 63.23 & 64.49 & 69.57 & 0.092 \\
      \bottomrule
    \end{tabular}%
  }
\end{table}

InteractVLM~\cite{dwivedi2025interactvlm} uses a VLM prior to predict an \texttt{<HCONTACT>} token, which is then decoded into human-mesh contact. In the original release, the model trained with body-part supervision was reported to outperform the variant trained without body-part supervision, and this supervised variant was the one publicly released. During our re-evaluation, we identified a bug in the baseline evaluation pipeline: ground-truth contacted body parts were inadvertently used at test time, which leaked supervision into evaluation. We confirmed this issue with the InteractVLM authors via email.

After correcting the evaluation pipeline, we observed that the model trained \emph{without} body-part supervision performs better than the model trained \emph{with} body-part supervision. Therefore, in Table~\ref{tab:supp_damon_results}, we report three InteractVLM variants: \textit{InteractVLM (+Part Sup., original eval)}, \textit{InteractVLM (+Part Sup., corrected eval)}, and \textit{InteractVLM-w/o-body-parts}. For all comparisons in the main paper and supplementary material, we use \textit{InteractVLM-w/o-body-parts} as the InteractVLM baseline.

\section{Method Details}
\label{sec:supp-method-details}

\subsection{Implementation Details}
\label{sec:supp-implementation-details}

\noindent We fine-tune the DETR detector~\cite{carion2020detr} on the MMHOI~\cite{kogashi2025mmhoi} and DAMON~\cite{tripathi2023deco} datasets.
After running the DETR detector, we discard detections with confidence below $0.2$, enumerate all candidate human--object pairs, and remove pairs whose human and object boxes have zero overlap (IoU $=0$). We initialize InteractionFormer from DINOv2-L~\cite{oquab2023dinov2} weights and fine-tune it jointly with the contact decoder and the contact-presence classifier.
The contact decoder is initialized from pretrained SAM weights~\cite{kirillov2023segment}; we keep the SAM image encoder frozen and fine-tune only the SAM decoder. We optimize all trainable components with AdamW.
We use an initial learning rate of $5\times 10^{-6}$ for DINOv2 parameters and $1\times 10^{-4}$ for all other trainable parameters. We train for $20$ epochs, including $4$ warm-up epochs with a start factor of $10^{-3}$. Unless stated otherwise, we use a batch size of $16$ and train on $4$ NVIDIA RTX A6000 GPUs. The query projection $\phi$ is implemented as a two-layer MLP that maps the concatenated DETR query features to $D_{\text{IF}}=1024$.
We set the loss weights to $\lambda_{\text{2dfocal}}=4.0$, $\lambda_{\text{dice}}=1.0$, $\lambda_{\text{3dfocal}}=4.0$, $\lambda_{\text{sp}}=0.01$, and $\lambda_{\text{cp}}=1.0$.

\subsection{Contact Decoder}
\label{sec:supp-decoder}

Given refined HO tokens $\mathbf{T}^{(L)}$, we predict a binary per-vertex contact mask $\mathbf{m}_{h,o}\in\{0,1\}^{N_v}$ for each candidate pair $(h,o)\in\mathcal{P}$. Directly decoding all 6,890 SMPL vertices is expensive.
Following InteractVLM~\cite{dwivedi2025interactvlm}, we instead decode HO tokens into multi-view 2D contact maps, where each map gives pixel-wise contact likelihood on a 2D projection of the SMPL~\cite{SMPL:2015} mesh. We then lift these maps back to the 3D SMPL surface.

\paragraph{Projecting SMPL Mesh to 2D.}
For each human instance $h$, we render its SMPL mesh (with vertices $\mathbf{V}_h\in\mathbb{R}^{N_v\times 3}$) in $V$ fixed viewpoints.
For view $v\in\{1,\dots,V\}$, we obtain a rendered image $\mathbf{I}^{(v)}_h$ and a dense correspondence between image pixels and mesh vertices.
We denote by $\pi_v:\{1,\dots,N_v\}\to\Omega_v\cup\{\varnothing\}$ the projection that maps a visible vertex index $j$ to its pixel location $\pi_v(j)\in\Omega_v$ (and to $\varnothing$ if the vertex is not visible in view $v$).

\paragraph{Decoding View-wise Contact Maps.}
For each view $v$, we feed the rendered image $\mathbf{I}^{(v)}_h$ through the SAM image encoder to obtain image features.
We then use the InteractionFormer-refined HO token $\mathbf{t}^{(L)}_{h,o}$ as the SAM object query and apply a shared SAM decoder head to predict a dense per-pixel contact probability map:
\begin{equation}
  \mathbf{P}^{(v)}_{h,o}=D_{\text{SAM}}\Bigl(E_{\text{SAM}}(\mathbf{I}^{(v)}_h),\,\mathbf{t}^{(L)}_{h,o}\Bigr)\in[0,1]^{|\Omega_v|},
\end{equation}
where $E_{\text{SAM}}(\cdot)$ and $D_{\text{SAM}}(\cdot)$ denote the SAM image encoder and mask decoder, respectively, and $\mathbf{t}^{(L)}_{h,o}$ is the refined HO token for the human--object pair $(h,o)$ after $L$ InteractionFormer refinement blocks.
We initialize $E_{\text{SAM}}$ and $D_{\text{SAM}}$ from pretrained SAM weights and fine-tune only the decoder for our task.

\paragraph{Lifting Contact Maps to a 3D Mesh.}
We lift the view-wise contact predictions back to the 3D human surface by pooling, for each vertex, the contact probabilities from all rendered views where that vertex is visible.
Let $\mathcal{V}_j=\{v\mid \pi_v(j)\neq\varnothing\}$ be the set of views where vertex $j$ is visible.
We compute the vertex contact probability as the mean of the corresponding per-view pixel probabilities:
\begin{equation}
  \hat{m}_{h,o}(j)=\frac{1}{|\mathcal{V}_j|}\sum_{v\in\mathcal{V}_j}\mathbf{P}^{(v)}_{h,o}(\pi_v(j))\in[0,1].
\end{equation}

\subsection{Training Objectives}
\label{sec:supp-training-objectives}

\paragraph{2D Supervision.}
Let $y^{(v)}_{h,o}(u)\in\{0,1\}$ be the ground-truth contact label for pixel $u\in\Omega_v$, defined only on valid rendered pixels $u\in\Omega_v^{\text{val}}$.
We supervise the view-wise predictions with a focal loss~\cite{lin2017focal} and a Dice loss~\cite{Sudre2017DiceLoss}:
\begin{equation}
  \mathcal{L}_{\text{2D}}=\lambda_{\text{2dfocal}}\,\mathcal{L}^{\text{focal}}_{\text{2D}}+\lambda_{\text{dice}}\,\mathcal{L}^{\text{dice}}_{\text{2D}}.
\end{equation}
\noindent where $\lambda_{\text{2dfocal}}$ and $\lambda_{\text{dice}}$ weight the focal and Dice losses, respectively.
Here, $\mathcal{L}^{\text{focal}}_{\text{2D}}$ denotes the standard focal loss and $\mathcal{L}^{\text{dice}}_{\text{2D}}$ denotes the standard Dice loss, computed between the predicted contact probabilities $\mathbf{P}^{(v)}_{h,o}$ and the 2D ground-truth masks $y^{(v)}_{h,o}$.

\paragraph{3D Supervision and Sparsity Loss.}
Let $m^{\ast}_{h,o}(j)\in\{0,1\}$ be the ground-truth contact label for vertex $j$.
We apply a focal loss on vertices and an $\ell_1$ sparsity penalty:
\begin{equation}
  \mathcal{L}_{\text{3D}}=\lambda_{\text{3dfocal}}\,\mathcal{L}^{\text{focal}}_{\text{3D}}+\lambda_{\text{sp}}\,\mathcal{L}_{\text{sp}},
\end{equation}
\noindent where $\lambda_{\text{3dfocal}}$ and $\lambda_{\text{sp}}$ weight the 3D focal loss and the sparsity regularizer, respectively.
We define the sparsity term $\mathcal{L}_{\text{sp}}$ as the $\ell_1$ penalty obtained by summing the predicted vertex contact probabilities $\hat{m}_{h,o}(j)$ over all vertices $j$ and all candidate pairs $(h,o)\in\mathcal{P}$.
Here, $\mathcal{L}^{\text{focal}}_{\text{3D}}$ denotes the standard focal loss between predicted vertex probabilities $\hat{m}_{h,o}(j)$ and ground-truth labels $m^{\ast}_{h,o}(j)$.

\paragraph{Contact Presence Supervision.}
We additionally supervise the contact presence classifier with a binary cross-entropy loss:
\begin{equation}
  \mathcal{L}_{\text{cp}}=\sum_{(h,o)\in\mathcal{P}} \mathrm{BCE}\bigl(p_{h,o},\, y_{h,o}^{\text{cp}}\bigr),
\end{equation}
where $y_{h,o}^{\text{cp}}\in\{0,1\}$ indicates whether the pair $(h,o)$ is in contact.

The total training loss is $\mathcal{L}=\mathcal{L}_{\text{2D}}+\mathcal{L}_{\text{3D}}+\lambda_{\text{cp}}\,\mathcal{L}_{\text{cp}}$.

\paragraph{Auxiliary Segmentation Losses.}
When ground-truth (GT) annotations are available, we further supervise the InteractionFormer image tokens with auxiliary semantic- and part-segmentation heads, following DECO~\cite{tripathi2023deco}.
Concretely, we reshape the patch tokens $\mathbf{X}^{(L)}$ into a 2D feature map and upsample it with a scene decoder $D_s$ and a part decoder $D_p$.
The scene decoder predicts a semantic segmentation map
\begin{equation}
  \bar{\mathbf{X}}_s = D_s(\mathbf{X}^{(L)}) \in \mathbb{R}^{H\times W\times N_o},
\end{equation}
where $N_o$ is the number of classes of MS-COCO objects.
Similarly, the part decoder predicts a human-part segmentation map
\begin{equation}
  \bar{\mathbf{X}}_p = D_p(\mathbf{X}^{(L)}) \in \mathbb{R}^{H\times W\times (J+1)},
\end{equation}
where $J$ is the number of body parts, and the extra channel corresponds to the background.
Both decoders are lightweight CNN heads implemented as a stack of transposed-convolution (ConvTranspose2D) layers that progressively upsample the patch-token feature map to the target resolution.
We denote the corresponding 2D segmentation losses (e.g., pixel-wise cross-entropy) between predictions and GT masks by $\mathcal{L}^{s}_{\text{2D}}$ and $\mathcal{L}^{p}_{\text{2D}}$.
These auxiliary terms are added to the training objective only when the GT segmentation masks are available:
\begin{equation}
  \mathcal{L}=\mathcal{L}_{\text{2D}}+\mathcal{L}_{\text{3D}}+\lambda_{\text{cp}}\,\mathcal{L}_{\text{cp}}+\mathbb{I}_{\mathrm{GT}}\Bigl(\lambda_s\,\mathcal{L}^{s}_{\text{2D}}+\lambda_p\,\mathcal{L}^{p}_{\text{2D}}\Bigr),
\end{equation}
where $\mathbb{I}_{\mathrm{GT}}\in\{0,1\}$ indicates whether GT segmentation supervision is present and $\lambda_s,\lambda_p$ are loss weights.

\subsection{Improving SAM3D Reconstruction}
\label{sec:supp-sam3d-refinement}

We initialize body pose and location from SAM3D Body~\cite{yang2026sam3dbody}, and object pose and mesh from SAM3D Object~\cite{sam3dteam2025sam3d3dfyimages}. We apply Pi-HOC to detect contact for every human--object pair in the scene. We transfer contacts from the SMPL mesh to the MHR mesh produced by SAM3D. Given the strong SAM3D initialization, we also derive object-side contact by mapping each human contact vertex to its nearest point on the object mesh.

Following InteractVLM~\cite{dwivedi2025interactvlm}, we optimize rotation $\mathbf{R}_h \in SO(3)$, translation $\mathbf{t}_h \in \mathbb{R}^3$, and scale $s_h \in \mathbb{R}^+$ for each human mesh $h$ to better satisfy contacts. At each iteration, we render the current human mesh using PyTorch3D and minimize the objective
\begin{equation}
  \mathcal{L} = \lambda_{\text{iou}} \mathcal{L}_{\text{iou}} + \lambda_{\text{cen}} \mathcal{L}_{\text{cen}} + \lambda_{\text{con}} \mathcal{L}_{\text{con}} ,
\end{equation}
where (1) $\mathcal{L}_{\text{iou}}$ is the mask IoU loss between the rendered mask and the ground-truth mask, (2) $\mathcal{L}_{\text{cen}} = \lVert \mu(\hat{M}) - \mu(M) \rVert_2$ aligns rendered and ground-truth mask centroids, and (3) $\mathcal{L}_{\text{con}}$ is the contact loss:
\begin{equation}
  \mathcal{L}_{\text{con}} =
  \frac{
    \sum_{i,j} p_i^h p_j^o
    \left\| \mathbf{v}_i^h - \mathbf{v}_j^o \right\|_2
  }{
    \sum_{i,j} p_i^h p_j^o
  }.
\end{equation}
Here, $\mathbf{v}_i^h$ and $\mathbf{v}_j^o$ are vertices on the human and object meshes, and $p_i^h, p_j^o \in \{0,1\}$ are their contact indicators. The product $p_i^h p_j^o$ gates each pair, so only vertices labeled as contact contribute to $\mathcal{L}_{\text{con}}$. We jointly optimize all human--object pairs.

\section{Semantic Contact Metrics - Object-Wise}
\label{sec:supp-objectwise-metrics}

\begin{table*}[t]
  \centering
  \caption{DAMON category-wise semantic contact results.}
  \label{tab:damon_categorywise_semantic_contact}
  \resizebox{\textwidth}{!}{%
    \begin{tabular}{lcccccccccccc}
      \toprule
      \multirow{2}{*}{Object Category}
      & \multicolumn{4}{c}{Semantic DECO~\cite{tripathi2023deco}}
      & \multicolumn{4}{c}{InteractVLM-w/o-body-parts~\cite{dwivedi2025interactvlm}}
      & \multicolumn{4}{c}{Pi-HOC (Ours)} \\
      \cmidrule(lr){2-5} \cmidrule(lr){6-9} \cmidrule(lr){10-13}
      & F1 (\%) $\uparrow$ & Prec (\%) $\uparrow$ & Rec (\%) $\uparrow$ & Geo (cm) $\downarrow$
      & F1 (\%) $\uparrow$ & Prec (\%) $\uparrow$ & Rec (\%) $\uparrow$ & Geo (cm) $\downarrow$
      & F1 (\%) $\uparrow$ & Prec (\%) $\uparrow$ & Rec (\%) $\uparrow$ & Geo (cm) $\downarrow$ \\
      \midrule
      Accessory
      & 40.1 & 30.7 & \textbf{75.6} & 21.88
      & \underline{55.44} & \underline{63.93} & 63.68 & \textbf{8.08}
      & \textbf{61.71} & \textbf{65.83} & \underline{69.35} & \underline{11.61} \\

      Daily Obj
      & 26.5 & 20.1 & 52.3 & 60.34
      & \underline{53.23} & \underline{53.86} & \underline{65.05} & \underline{10.96}
      & \textbf{58.00} & \textbf{57.85} & \textbf{69.43} & \textbf{10.66} \\

      Food
      & 11.7 & 19.4 & 12.9 & 49.61
      & \underline{57.12} & \underline{54.54} & \underline{69.72} & \textbf{9.80}
      & \textbf{58.06} & \textbf{56.45} & \textbf{73.98} & \underline{11.29} \\

      Furniture
      & 24.5 & 15.8 & \textbf{83.7} & 29.17
      & \underline{51.09} & \textbf{53.80} & 49.50 & \textbf{4.49}
      & \textbf{53.57} & \underline{52.78} & \underline{52.96} & \underline{6.07} \\

      Kitchen
      & 27.7 & 24.7 & 37.2 & 52.34
      & \underline{57.07} & \underline{55.81} & \underline{67.00} & \underline{23.79}
      & \textbf{63.32} & \textbf{62.82} & \textbf{69.78} & \textbf{22.98} \\

      Sports
      & 36.4 & 30.4 & \textbf{80.1} & 79.21
      & \underline{65.57} & \underline{56.26} & \underline{71.51} & \underline{10.49}
      & \textbf{68.29} & \textbf{66.30} & 68.62 & \textbf{10.30} \\

      Transport
      & 52.0 & 39.1 & \textbf{93.7} & 31.78
      & \underline{68.22} & \underline{67.83} & 75.80 & \underline{4.07}
      & \textbf{70.86} & \textbf{69.51} & \underline{80.77} & \textbf{2.87} \\
      \bottomrule
    \end{tabular}
  }
\end{table*}

Table~\ref{tab:damon_categorywise_semantic_contact} shows that Pi-HOC is robust across object categories and generally achieves higher contact precision with lower geodesic error, indicating more fine-grained and accurate contact localization.

\section{Qualitative Results}
\label{sec:supp-qualitative}

\begin{figure*}[t]
  \centering
  \includegraphics[width=\textwidth,height=0.92\textheight,keepaspectratio]{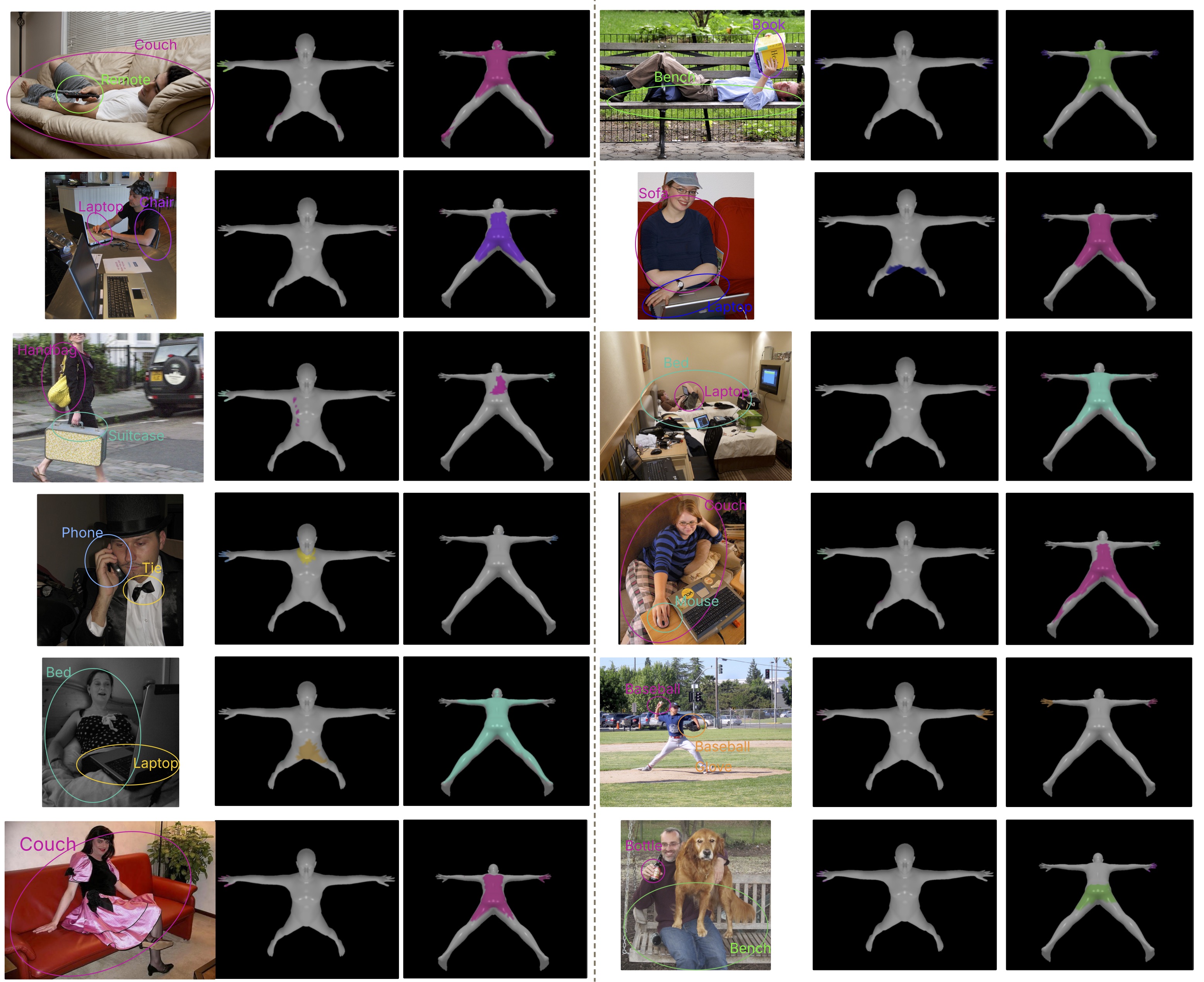}
  \caption{\textbf{Semantic Human Contact Estimation. }
  We show semantic contact estimation on in-the-wild images. Each example shows all objects with which the human is in contact and the corresponding contact regions on the human mesh.}
  \label{fig:supp_semantic_contact}
\end{figure*}

\begin{figure*}[t]
  \centering
  \includegraphics[width=\textwidth,height=0.95\textheight,keepaspectratio]{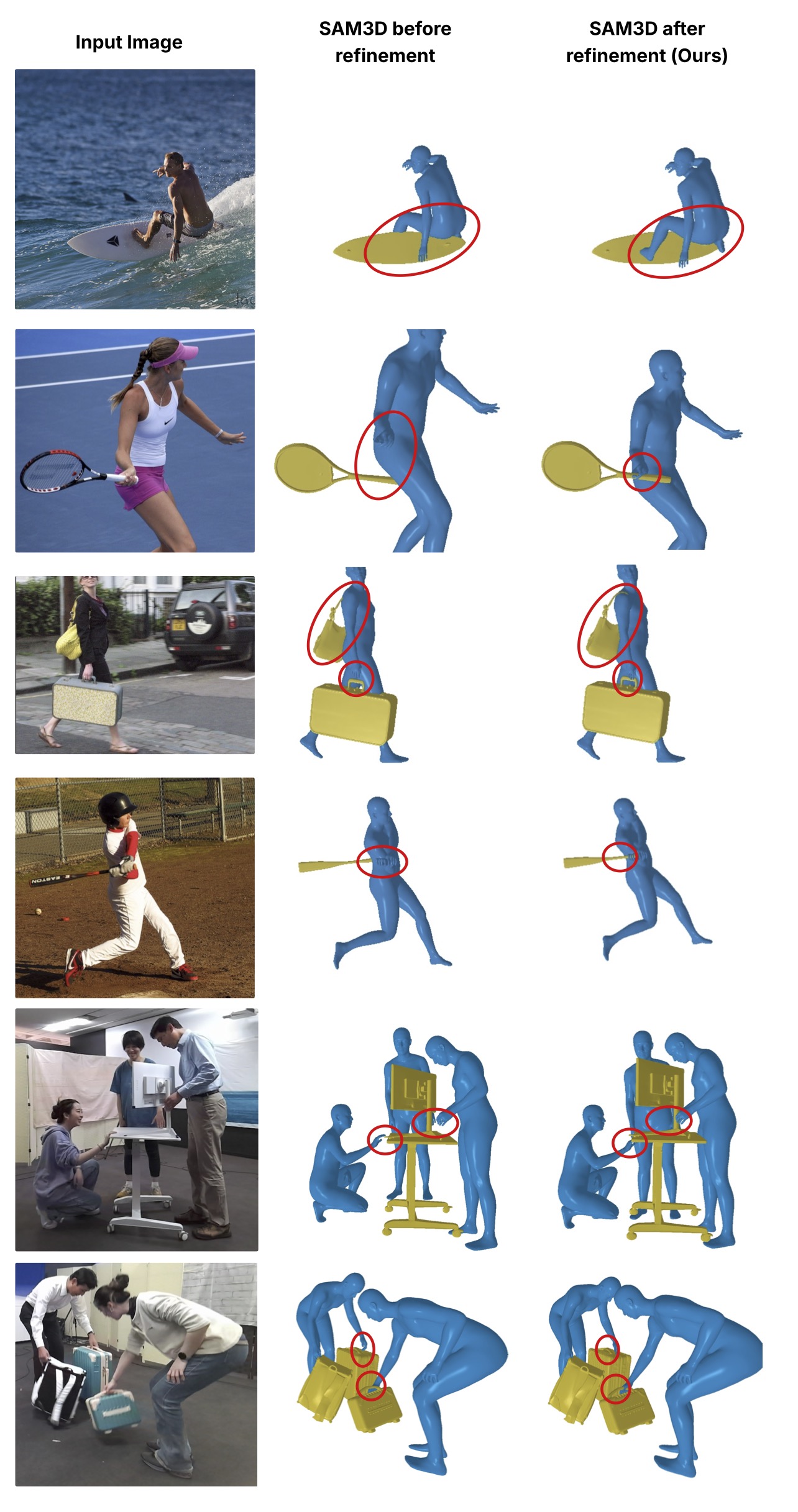}
  \caption{\textbf{SAM3D Test-Time Refinement.}}
  \label{fig:supp_sam3d_refinement}
\end{figure*}

\begin{figure*}[t]
  \centering
  \includegraphics[width=\textwidth,height=0.95\textheight,keepaspectratio]{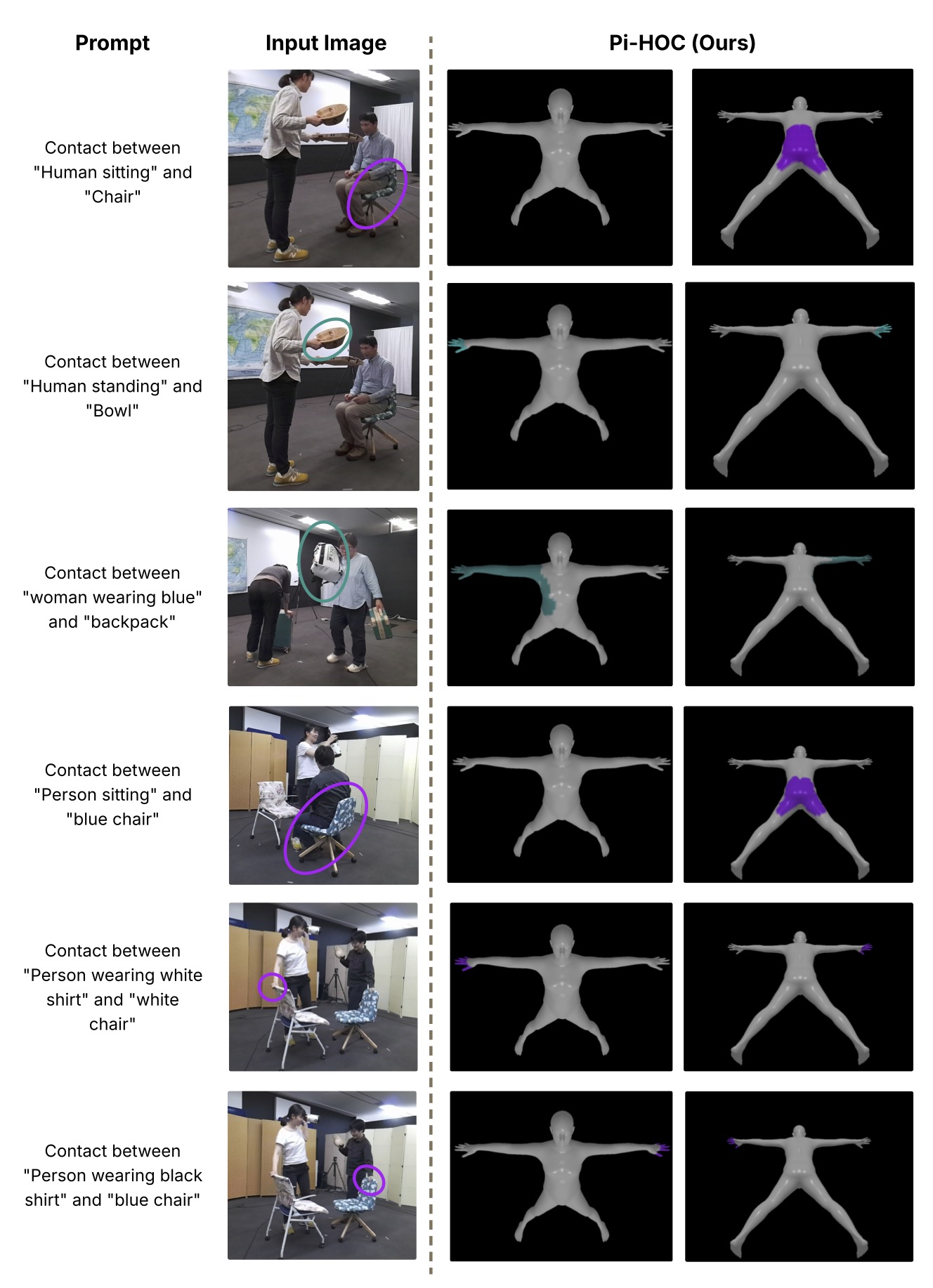}
  \caption{\textbf{Pi-HOC Referential Contact.}}
  \label{fig:supp_referential_contact}
\end{figure*}

We present qualitative results of our method. As shown in Fig.~\ref{fig:supp_semantic_contact}, it successfully detects contacting human--object pairs and colors the human mesh according to object-specific contact regions on the body surface. Compared with prior methods~\cite{dwivedi2025interactvlm}, which predict contacts one human--object pair at a time, our method predicts all contacts in a single forward pass.

We also present additional results on SAM3D test-time refinement. We use the contact predictions from our model to optimize the SAM3D-initialized reconstructions. As shown in Fig.~\ref{fig:supp_sam3d_refinement}, the method removes artifacts such as floating hands and produces more physically plausible reconstructions.

We also showcase that our model can be adapted to referential contact estimation without additional training. As shown in Fig.~\ref{fig:supp_referential_contact}, the method predicts contact only for the human and object instances specified by the text prompt.

\section{Failure Cases}
\label{sec:supp-failure-cases}

\begin{figure*}[t]
  \centering
  \includegraphics[width=\textwidth,height=0.92\textheight,keepaspectratio]{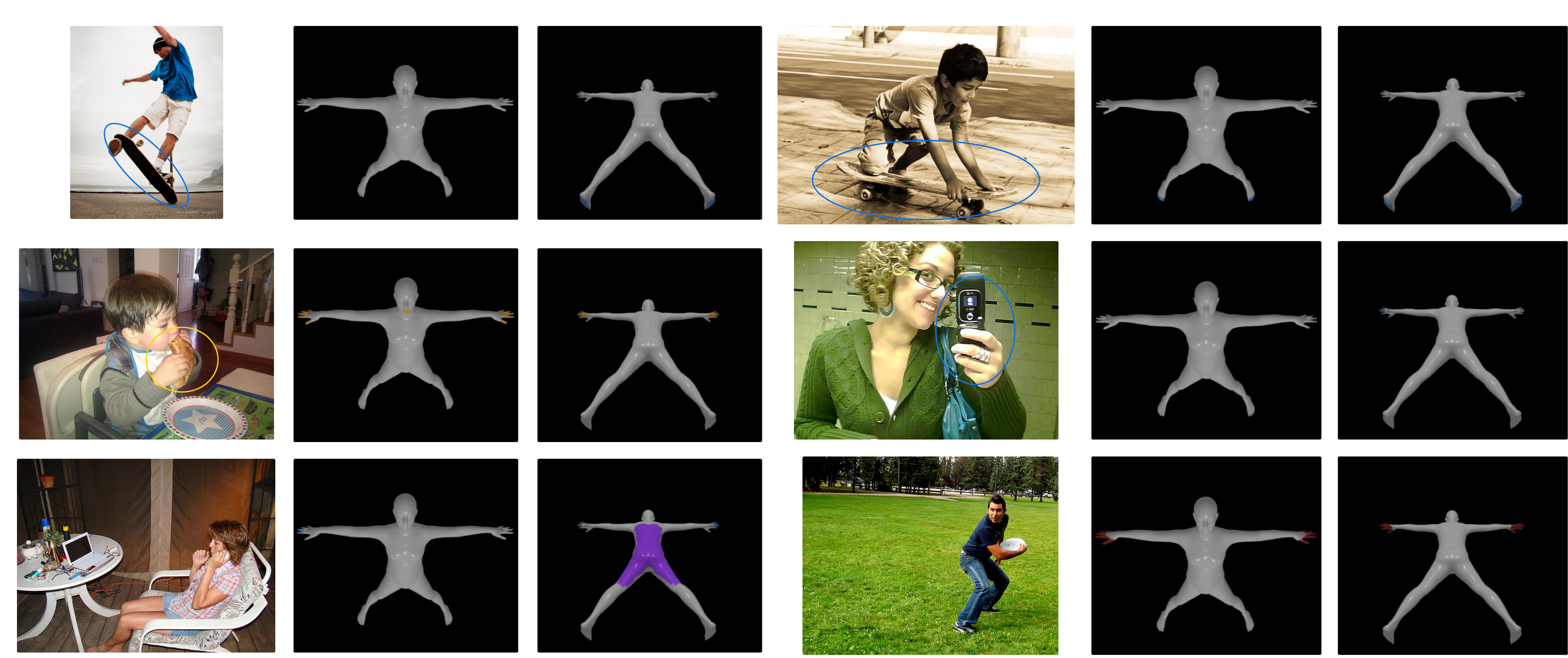}
  \caption{\textbf{Semantic Human Contact Failure Cases. }
  Each row shows one failure case: (1) interactions that deviate strongly from common interaction patterns, (2) missed DETR detections that prevent human--object pair formation (e.g., row 2: the child--chair interaction in the left image and the woman--handbag contact on the right image), and (3) confusion between left and right hands when they are very close.}
  \label{fig:supp_failure_case}
\end{figure*}

\begin{figure*}[t]
  \centering
  \includegraphics[width=\textwidth,height=0.92\textheight,keepaspectratio]{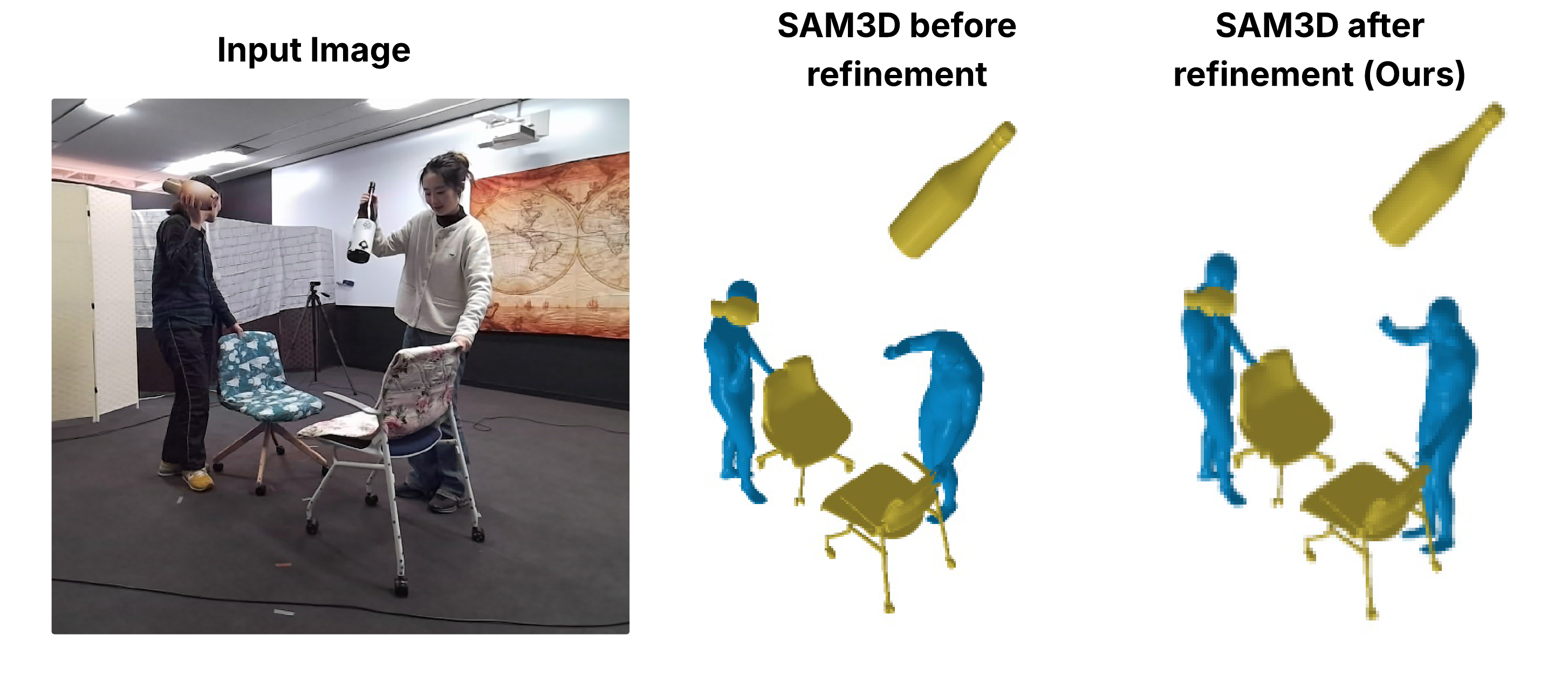}
  \caption{\textbf{SAM3D Refinement Failure Cases. }
  If the SAM3D initialization is severely misaligned, the algorithm can fail because we keep the object fixed and optimize only the human rotation, translation, and scale.}
  \label{fig:supp_sam3d_reconstruction_failure_case}
\end{figure*}

Despite strong performance, our method fails in certain cases, as shown in Fig.~\ref{fig:supp_failure_case}. We observe three common failure scenarios: (1) interactions that differ significantly from common interaction patterns, (2) missed DETR detections that prevent pair formation and therefore contact estimation, and (3) ambiguity between left and right hands when they are close together, which can lead to incorrect contact localization. More robust DETR fine-tuning, or the use of a stronger detector, could help reduce these errors.

For the test-time SAM3D refinement algorithm, failures occur when the SAM3D-initialized reconstruction is severely misaligned and optimization cannot recover a plausible reconstruction. This happens because we keep the object position fixed and optimize only the human mesh rotation, translation, and scale. If the object placement in the initial reconstruction is incorrect, refinement often fails (Fig.~\ref{fig:supp_sam3d_reconstruction_failure_case}).

\section{Attention Visualizations}
\label{sec:supp-attn}

\begin{figure*}[t]
  \centering
  \includegraphics[width=\textwidth,height=0.85\textheight,keepaspectratio]{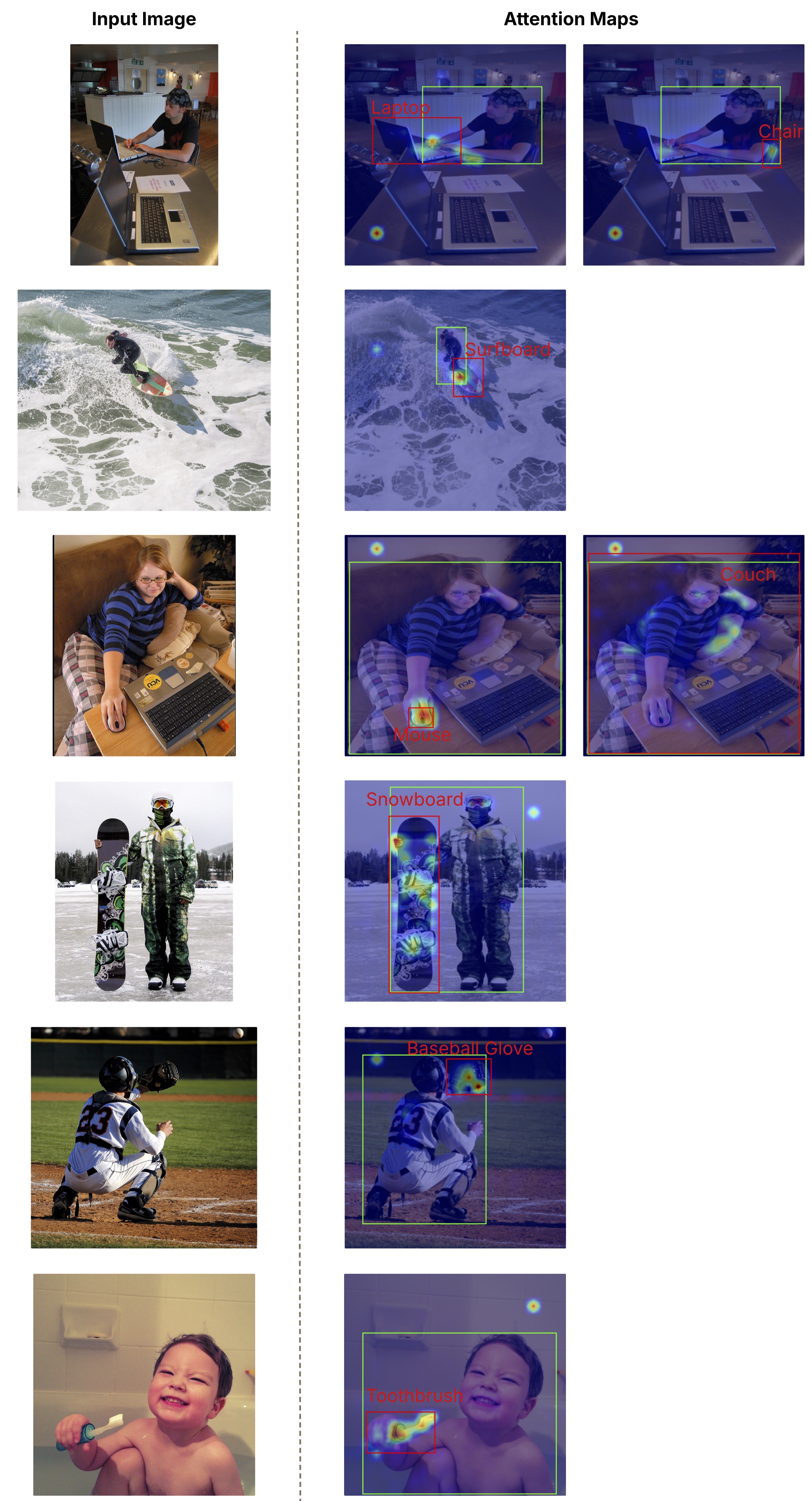}
  \caption{\textbf{InteractionFormer Attention Maps.} Each row starts with the input image, followed by attention maps for individual contacting human--object pairs. The blue box denotes the human instance and the red box denotes the corresponding object instance. The visualizations show that each pair token focuses primarily on interaction-relevant body and object regions, while also attending to limited contextual cues.}
  \label{fig:supp_attn_maps}
\end{figure*}

We visualize attention maps from the last InteractionFormer layer for each contacting human--object pair in an image. As shown in Fig.~\ref{fig:supp_attn_maps}, each pair token concentrates on spatially localized regions that are most informative for contact reasoning, such as hands, forearms, and the nearby object surface. This behavior provides useful intuition for why the model produces precise contact predictions. Rather than relying on global scene context alone, the token learns interaction-specific correspondences between human parts and object regions.

\end{document}